\documentclass[10pt,letterpaper]{article}

\usepackage{cogsci}
\usepackage{pslatex}
\usepackage{apacite}
\usepackage{graphicx}
\usepackage{gensymb}
\usepackage{afterpage}
\usepackage{amsfonts}
\usepackage{subfig}
\usepackage{natbib}
\usepackage{caption}
\usepackage{subfloat}

\title{Structure Learning in Motor Control: \\A Deep Reinforcement Learning Model}
 
\author{{\large \bf Ari Weinstein (ariweinstein@google.com)} \\ 
  Deepmind, London, UK\\
  \AND {\large \bf Matthew M. Botvinick (botvinick@google.com)} \\
  Deepmind, London, UK\\
  Gatsby Computational Neuroscience Unit, University College London\\}

\begin{document}

\maketitle

\begin{abstract}
Motor adaptation displays a structure-learning effect: adaptation to a new perturbation occurs more quickly when the subject has prior exposure to perturbations with related structure. Although this `learning-to-learn' effect is well documented, its underlying computational mechanisms are poorly understood. We present a new model of motor structure learning, approaching it from the point of view of deep reinforcement learning. Previous work outside of motor control has shown how recurrent neural networks can account for learning-to-learn effects. We leverage this insight to address motor learning, by importing it into the setting of model-based reinforcement learning. We apply the resulting processing architecture to empirical findings from a landmark study of structure learning in target-directed reaching \citep{BraunAW09}, and discuss its implications for a wider range of learning-to-learn phenomena. 

\end{abstract}

\section{Introduction}
Learning can be defined as a process that improves performance as exposure to a task increases. However, research on human and animal learning makes clear that this simple definition is not quite enough to explain the observed relationship between experience and performance. The full picture must also include `learning-to-learn,' a process whereby growing experience causes learning itself to become more efficient~\citep{Harlow49}. More specifically, learning-to-learn (also referred to as meta-learning and structure learning) occurs in settings where the learner encounters a series of tasks that share some underlying structure, and gains from these an ability to quickly adapt to a new task that displays the same general form~\citep{ThrunL98}. 

A vivid example of learning-to-learn, which provides a concrete focus for the present research, comes from research on motor adaptation. Many studies have documented the ability of human subjects to adapt to perturbations of motor dynamics or kinematics, as for example in prism adaptation~\citep{Harris63}. However, a series of studies by Braun and colleagues~\citep{BraunAW09, BraunMW10, Braun2012} went beyond this to show that adaptation can occur faster when the subject has prior exposure to perturbations that share structure with the final test conditions. In one specific experiment, upon which we will continue to focus, Braun and colleagues~\citeyearpar{BraunAW09} studied reaching under visuomotor rotation. They examined the speed with which target-directed reaching adapted to a 60-degree rotation, manipulating between subjects the content of a preceding set of training trials. In one condition, which we will refer to as \textit{Rot}, subjects dealt with a series of rotations (though never the one presented at test). In a comparison condition \textit{Rot+}, subjects dealt with a more diverse set of transformations, each made up of a rotation along with shear and scale components. Results showed that subjects in the \textit{Rot} group adapted faster to the probe rotation problem (Figure~\ref{fig:braunfig}). Braun et al.~\citeyearpar{BraunAW09} interpreted this as learning-to-learn effect, which they referred to as ``motor structure learning'': Subjects in the \textit{Rot} group evidently learned that the transformations being presented were restricted to a particular structurally coherent set (rotations), and this allowed them to infer and adapt rapidly to the probe transformation. This structure learning was less feasible in the \textit{Rot+} condition because the structure underlying the training set was more complex, thus offering weaker constraint on inference when facing a new transformation. 

\afterpage{%
    \begin{figure}
    \begin{center}
    \includegraphics[width=0.33\textwidth]{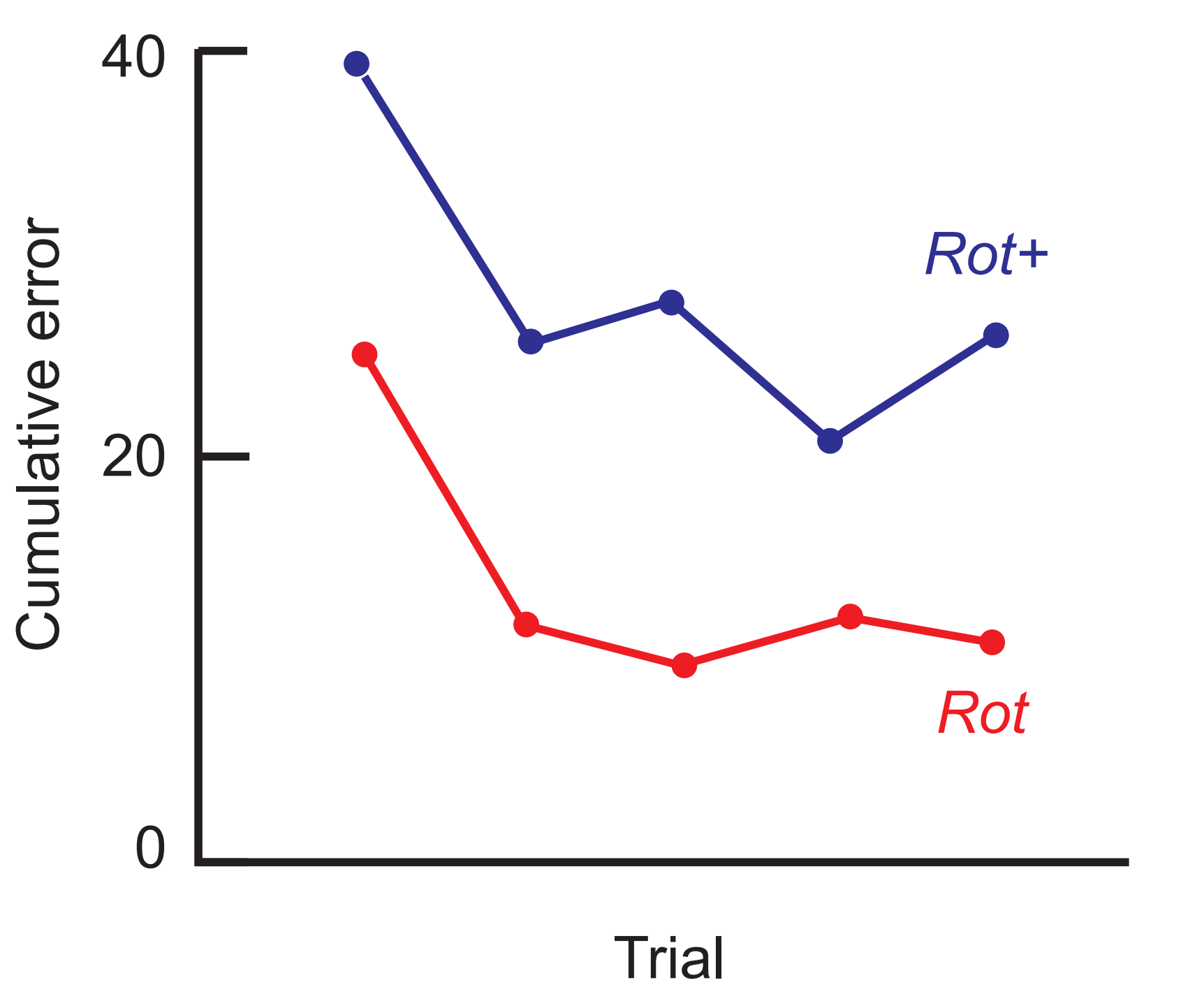}
    \end{center}
    \caption{Results redrawn from Braun et al.(2009), showing mean cumulative error on a series of five reaches under a 60-degree visuomotor rotation.}
    \label{fig:braunfig}
    \end{figure}
}

In the present study, we consider the computational mechanisms underlying motor structure learning, treating it as a case study in learning-to-learn. Despite widespread agreement that learning-to-learn effects are both real and important, the precise computational processes underlying such effects are poorly understood. The most widely proposed idea comes from a Bayesian perspective, and proposes that learning-to-learn involves refining the structure and hyperparameters of a generative model of the relevant task domain~\citep{LakeST15}. Braun and colleagues initially  proposed, and later investigated~\citep{GeneweinEtAl15} a model of this sort to account for their structure learning results. 

A different computational proposal, which has been less widely considered in cognitive science, comes from neural network or deep learning research. In classic work, Hochreiter and colleagues~\citeyearpar{HochreiterYC01} showed how a recurrent neural network (RNN) can learn to learn, by integrating information about past outcomes into predictions concerning new observations. Recent applications of this idea~\citep{WangEtAl16, DuanEtAl16} have treated the RNN in Hochreiter's scheme as a mechanism for directly selecting actions. In the present work, we leverage Hochreiter's~\citeyearpar{HochreiterYC01} insight in a different way, using an RNN as an adaptive model of the task domain, which is leveraged by a separate action-selection mechanism. In this sense, the aim of our work is to bridge Bayesian and deep learning perspectives on learning-to-learn.  On a more immediate level, we show how the resulting approach can be used to account for the findings of Braun and colleagues~\citeyearpar{BraunAW09} in motor adaptation.

The motor control literature suggests that actions such as reaching are based, at least in part, on an internal predictive or forward model of reaching dynamics~\citep{WolpertGJ95, MiallW96}, and analyses of motor adaptation have portrayed adaptation as reflecting a progressive adjustment of this internal model to fit with current observations~\citep{BernikerK08,HaithK13}.  Following this idea, and a great deal of previous work in computational motor control, we construe action selection as a form of model-based reinforcement learning (RL). 

Formulating the problem in this way begins by casting it as a finite-time Markov decision process (MDP) $M$, which is made of a set of states $S$, a set of possible actions $A$, a transition function $\mathcal{T}$, and reward function $R$ (in many settings a discount factor $\gamma$ is included, but since we formulate the task as a finite-time problem this is unnecessary).  The goal is to select actions that maximize the cumulative reward up to some time $T$: $\sum_{t=0}^T r_{t+1}$, where $t$ indexes discrete time steps up to some maximum $T$, and $r_t$ is the reward received on each step.
Focusing on target-directed reaching, the task studied by Braun and colleagues~\citeyearpar{BraunAW09}, the problem is defined as follows: $M$ is the entire reaching task; $S$ are possible arm configurations; $A$ are possible motor inputs; $\mathcal{T}$ defines the dynamics of the arm based on motor inputs; $R$ is the negative distance from the cursor to the target.  In order to be consistent with the literature on structure learning in motor control, we will use the terms reward maximization and penalty (or error) minimization interchangeably.

\begin{figure}[t]
\begin{center}
\includegraphics[width=\linewidth]{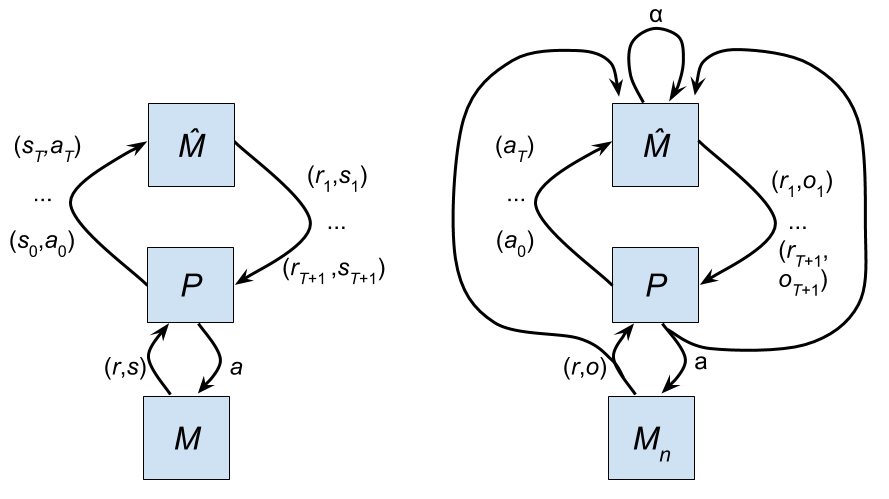}
\end{center}
\caption{Model-based reinforcement learning with a fixed model (left) and an adaptive model (right).} 
\label{fig:diagram}
\end{figure}

In model-based RL, a model $\hat{M}$ of the environment $M$ is built, and then used by a planner $P$ in order to construct an action-selection policy. The general form of a model-based learning architecture is diagrammed in Figure~\ref{fig:diagram}, left.  Here a planner $P$ is informed of a current state $s$ by the true MDP $M$.  Based on the particular policy of $P$, the planner queries the model $\hat{M}$ with a series of state-action pairs $(s_t, a_t)$, and in turn receives an estimated next state $s_{t+1}$ and reward $r_{t+1}$.  After the planner completes querying $\hat{M}$, either because it has taken as much data as it needs or due to some outside pressure such as a time limit, it returns an action $a$ which is executed in $M$, which results in a new state and reward and the process repeats. 

In the architecture shown in Figure~\ref{fig:diagram} (left), one way of implementing the forward model $M$ is as a feed-forward neural network. This approach has been explored in a number of previous studies~\citep{JordanR92, HamrickEtAl16}. However, a feed-forward neural network will not suffice to address the learning-to-learn phenomena we are concerned with here. Indeed, the overall architecture must be fundamentally changed in order to address the learning-to-learn problem. 

As introduced earlier, learning-to-learn arises in a setting where the learner encounters a series of interrelated problems or tasks, and must adapt to each one in turn. Using our terminology, each task $M_n, n=1...N$ becomes a sample from a task distribution $\mathbb{M}$.  As such, the properties of each $M_n$ must be inferred based on observed action-outcome pairs (a process referred to in the engineering literature as \textit{system identification}). On a formal level, this demand changes the MDP we have been considering into a partially observable Markov decision process (POMDP). By definition, a POMDP is an MDP which additionally has an observation space $O$ and observation function $\Omega$ which takes its internal state and outputs an observation $o$ to the agent. Instead of the true state, an agent only has access to observations, which unlike state, is generally insufficient to act optimally when considered in isolation. In order for $\hat{M}$ to adjust to each $M_n$, it must have some form of memory $\alpha$ to keep a relevant summary of interactions with the environment, allowing for integration over previous timesteps in order to accurately estimate problem dynamics. 

These requirements yield the interaction and planning structure diagrammed in Figure~\ref{fig:diagram} (right).  Instead of states $s$, information presented is in terms of observations $o$. The model $\hat{M}$ must now directly consume $o$ and $r$ from $M_n$ at every time step, which causes it to update its memory $\alpha$.  $P$ now only passes a sequence of actions along trajectories to the model.  This is because it does not have access to the true state, and because observations alone are not sufficient for planning or modelling\footnote{Belief states~\citep{KaelblingLC98} or predictive state representations~\citep{LittmanSS01} are sufficient for planning, but can be computed internally in each module and do not need to be communicated.}.  Additionally, during the planning trajectories, the $P$ must signal to $\hat{M}$ when its evaluation of a simulated trajectory is complete so that $\alpha$ can be reset (omitted from figure for clarity).

Note that, unlike in the simpler MDP, in the POMDP setting the internal model $\hat{M}$ cannot be accurately implemented as a feed-forward neural network, because such networks do not have memory or persistent internal state. The key move in the present work is to substitute for the feed-forward network a RNN, whose recurrent connectivity endows it with the memory needed to support system identification and, as we will show, learning-to-learn. 

\section{Simulation Study}

We predicted that the proposed architecture, if trained in an appropriate multi-task setting, would display learning-to-learn, leveraging experience with past tasks to adapt rapidly to a new task sharing in the same structure. In order to test this idea, we applied the architecture to the task paradigm employed in they study of motor structure learning by Braun and colleagues~\citeyearpar{BraunAW09}.

\subsection{Model implementation and task design}

We implement the architecture shown in Figure~\ref{fig:diagram} (right), instantiating the forward model in the form of a recurrent neural network (which is naturally deep as it is unrolled over time). More specifically, this involves one LSTM layer~\citep{HochreiterS97} followed by two more fully connected layers containing rectified linear units~\citep{NairH10}, where each layer contains 100 units.  The planner is an open-loop planner based on cross-entropy optimization, as described in~\citeauthor{WeinsteinL13}~\citeyearpar{WeinsteinL13}, with the addition of ``warm starting.'' In warm starting, planning is done from scratch on the first step of a trajectory, but all subsequent steps in the actual domain initiate planning with the result from the previous step. At each time step only the first action in the current plan is executed in the true domain before partial replanning in this manner occurs.  For simplicity, we assume (without loss of generality) that $\hat{M}$ has access to the reach-target coordinates and can compute the reward function. 

In order to model target-directed reaching, we implemented a simple arm model.  While not intended to offer a detailed model of biomechanics, this was intended to capture the most important aspects in terms of possible arm geometry, velocity, and acceleration~\citep{Nagasaki89}.  As simulated, the underlying state space of the problem has four dimensions: horizontal shoulder angle, elbow angle, and corresponding angular velocities. Observations emitted are the Euclidian position of the cursor controlled by the arm's tip as seen in the experiment (meaning that $\hat{M}$ must also learn to estimate velocities), and the goal.  The two dimensional action space sets the angular accelerations of the joints, and the reward is the negative Euclidean distance of the cursor from the center of the goal region.  

In the reaching task, the cursor is always initialized at the origin and is controlled by the transformation of the underlying position of the simulated hand.  Before each trial a goal location is selected which is set to be 8 cm from the origin at a uniformly distributed angle.  


\subsection{Training and testing procedure}

\begin{figure*}[t]
\centering
\minipage{0.35\textwidth}
  \includegraphics[width=\linewidth] {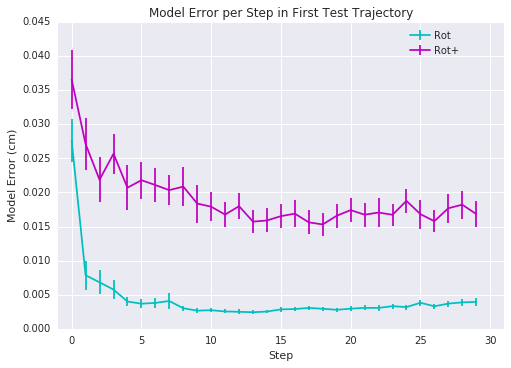}
  \caption{Model errors by step in initial trial.}
  \label{fig:modelerror}
\endminipage
\hspace{0.1\linewidth}
\quad
\minipage{0.35\textwidth}%
  \includegraphics[width=\linewidth] {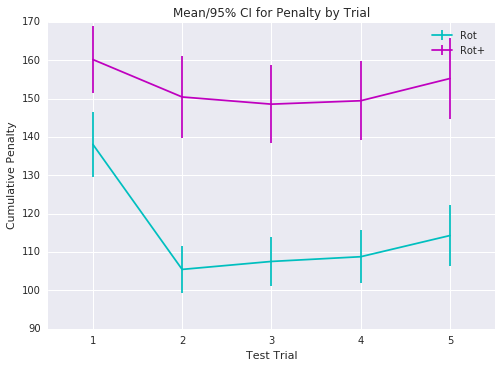}
  \caption{Average cumulative penalty by trial.}
  \label{fig:rewards}
\endminipage
\end{figure*}

The simulation study, like the experiment by~\citeauthor{BraunAW09}~\citeyearpar{BraunAW09} was divided into training and testing phases.  During training, the RNN model was trained to predict each sequential outcome observation exactly along a trajectory consisting of observations and randomly selected actions, that is, following a random walk. Again as in the empirical study, two versions of the model were trained in different  environments. One model, which we label (in a minor abuse of terminology) \textit{Rot}, was trained on a series of visuomotor rotations, simulated by appropriately transforming the observed cursor coordinates. The second model instance, \textit{Rot+}, was trained on a combination of rotations, shears, and scales (following the design described in~\citeauthor{BraunAW09},~\citeyear{BraunAW09}, Supplemental Data). Following the design imposed by Braun and colleagues, when the rotation to be presented to \textit{Rot+} fell between $\pm 50\degree$ and $70\degree$, a rotation of $\pm 60\degree$ was substituted and no linear transform was applied. As a result, both \textit{Rot} and \textit{Rot+} had roughly equal exposure to the transformation used during test trials. In both conditions, the model was trained by backpropagation through time on 2,000 trajectories of random-walk data, with each trajectory containing three seconds of simulation time, and training starting from the initial observation of each trajectory.  

In the testing phase the RNN weight parameters were frozen and reaches were elicited only under only pure rotations of $\pm 60\degree$, as in the testing phase of the experiment by Braun and colleagues. Goal locations were placed at a randomly selected angle 8 cm from the start location of the cursor. The radius of the goal region is 1.6 cm. In order to simulate a series of reaches, the angle of the imposed visuomotor rotation was held constant while the position of the goal varied between reaches. Test reach trajectories ran for a maximum of two seconds, terminating early if the cursor was brought within the goal region for 500 ms. 

\subsection{Results}

Training of models for both the \textit{Rot} and \textit{Rot+} conditions were successful, but the model trained on \textit{Rot} was able to achieve an average error of about $0.002$ cm per time step for trajectories in the training set, while \textit{Rot+} an error of about $0.03$ cm by the same metric.  In the \textit{Rot} condition, the RNN model learned to act as an adaptive forward model, adjusting its predictions to fit with accumulating action-outcome observations. Figure~\ref{fig:modelerror} shows the average observation-prediction errors of both \textit{Rot} and \textit{Rot+} models during an initial random-walk trajectory which was not part of the training set. The initial data-point is the error of the model prior to any experience in the test MDP.  In interpreting the values on the y-axis of the plot, it should be taken into account that in our simulation two seconds of time takes 28 discrete time steps, and error compounds over these steps. 
In contrast to the \textit{Rot} model, the \textit{Rot+} model adapts much less successfully, despite having been trained on an identical amount of data. 

Figure~\ref{fig:rewards} shows the mean cumulative penalty when the model is coupled with a planner, for each reach at test for both models.  This is intended for comparison with the empirical data from humans in preceeding work shown in Figure~\ref{fig:braunfig}. As predicted, \textit{Rot} is better able to conduct structure learning, by adapting more rapidly and completely to the test rotation (the manipulation both models were exposed to during training) than \textit{Rot+}.  This qualitatively replicates the experimental findings from Braun et al.~\citeyearpar{BraunAW09}.

Figure~\ref{fig:avgtrajectories} shows average trajectories for five successive reaches (normalized by rotation and goal angles), for both \textit{Rot} and \textit{Rot+} models.  Both models adapted across reaches (starting with smaller initial angular errors after the initial reach), but the effects were stronger in the \textit{Rot} model.  Quite striking is the standard deviation of the final position of the first trial of \textit{Rot}, and \textit{Rot+} in cyan and magenta, respectively.  Although on average \textit{Rot+} tracked toward target, there is a tremendous amount of variability in its trajectories, and was not able to consistently reach the goal region, whereas \textit{Rot} usually terminated within the target.

We also consider other indirect metrics of performance which are presented in the human studies such as initial angular error, velocity, and minimum distance to goal region, which are presented in Figures~\ref{fig:angleerrors} through \ref{fig:distogoal}, respectively.  In general the results with these metrics are similar to the previous plots, with \textit{Rot} improving quickly and performing better than \textit{Rot+}.  We also note the higher variance of \textit{Rot+}, which manifests itself in wider confidence intervals across all Figures, especially Figure~\ref{fig:distogoal}.  These results are qualitatively aligned with those reported in the experimental study.

In fact, of these metrics, the only one which shows improvement by \textit{Rot+} is the initial angular error.  Even with this improvement, the agent frequently falls short of reaching goal region (which would allow for an early termination of distance penalties).  This is most likely due to the fact that on average testing data in \textit{Rot+} has a scaling amount of roughly 1.3 (this design is part of the original human study), and indeed \textit{Rot+} almost uniformly tells the planner that actions will result in greater changes in location than actually occur.  

\begin{subfigures}
\begin{figure*}[t]
\centering

\minipage{0.35\textwidth}
  \includegraphics[width=\linewidth] {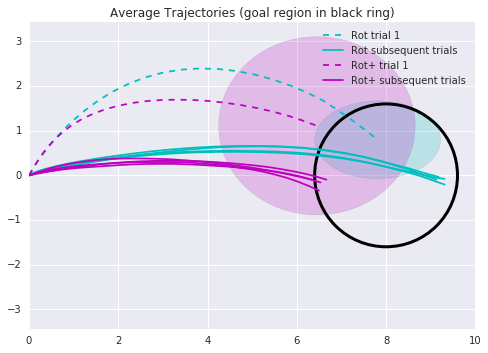}
  \caption{Average trajectories by trial.  Standard deviation of final position of first trial in shaded region.  Goal region in black.}
  \label{fig:avgtrajectories}
\endminipage
\hspace{0.1\linewidth}
\quad
\minipage{0.35\textwidth}%
  \includegraphics[width=\linewidth] {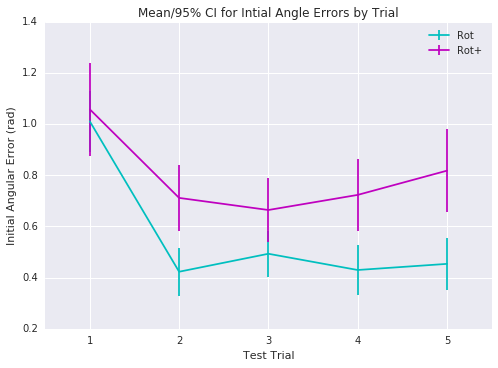}
  \caption{Average angle error from goal after 200 ms of simulated time.}
  \label{fig:angleerrors}
\endminipage
\vspace{5 mm}

\minipage{0.35\textwidth}
  \includegraphics[width=\linewidth] {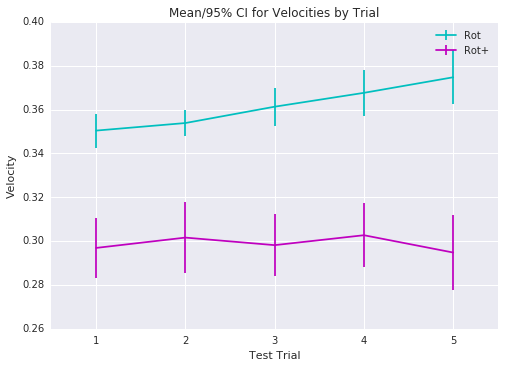}
  \caption{Average velocities by trial}
  \label{fig:velocities}
\endminipage
\hspace{0.1\linewidth}
\quad
\minipage{0.35\textwidth}%
  \includegraphics[width=\linewidth] {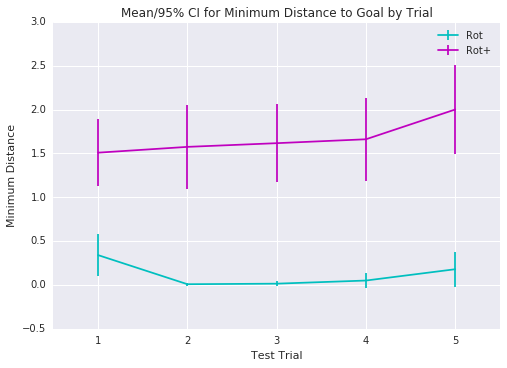}
  \caption{Average minimum distance to goal by trial}
  \label{fig:distogoal}
\endminipage
\end{figure*}
\end{subfigures}

Although \textit{Rot+} was less effective at structure learning than \textit{Rot}, it is not the case that it failed entirely.  The average penalty of a trajectory for agent using a uniform random policy is approximately 220 units which is significantly poorer than what \textit{Rot+} was able to achieve.  
 
We note that our goal was not to fit the results from the human data quantitatively, but rather to demonstrate the same phenomenon which is that structure learning becomes more difficult as the the amount of variability in the problem increases.  And although \textit{Rot+} was not able to perform well, the overall architecture does have the capacity to do effective structure learning; expanding the data corpus size by a factor of five produces models that have statistically equal, high quality performance on test tasks for both \textit{Rot} and \textit{Rot+}.  

\section{Discussion}

Learning-to-learn is a fundamental aspect of human behavior, but its computational basis is not yet well understood. We have presented a new model of learning-to-learn in the setting of motor adaptation. This task, defined by Braun and colleagues~\citeyearpar{BraunAW09} involves learning to learn in the sense that the subject must gather data on a current situation in order to infer the hidden parameters of the dynamics, and indeed Braun and colleagues state that learning to learn can be recast as structure learning. On the other hand, a stronger definition of learning to learn could require learning to adapt to a situation it has not experienced in the past, perhaps in terms of new objects to interact with that follow some prelearned rules~\citep{Harlow49}.  This has been considered in a different simulated setting in RL where the agent learns policies~\citep{WangEtAl16}, as opposed to models of the environment as is done here.

Adopting the standard approach, we assume that motor adaptation involves updating an internal forward model of reaching dynamics. Our novel contribution is to instantiate this internal model as a recurrent neural network. Through simulations of a key experimental study, we have shown that the resulting system not only learns to adapt to changing perturbations, but also that its adaptation becomes more effective when there is prior exposure to structurally related conditions, as seen empirically in motor structure learning. Importantly, no special measures were required in order to secure this learning-to-learn effect. Through error-correcting learning, the parameters of the RNN are, perforce, fit to the structure of the pre-training data. That same structure is thus naturally -- indeed inevitably -- expressed in its later inferences at test. 

We consider learning to learn as refining a (potentially implicit) hypothesis set based on experience.  If the problem has a large underlying dimension, then the hypothesis set learned by the model must be of corresponding size.  This is in turn fundamentally linked to the amount of data required to both train the model, as well as do inference, accurately. For these reasons, it is to be expected that when comparing the data requirements of doing both in \textit{Rot} versus \textit{Rot+}, \textit{Rot} leads to lower data requirements. Just as is the case with Braun and colleagues~\citeyearpar{BraunAW09}, we do not attempt to disentangle these issues, although a more detailed investigation warrants future attention.

As noted earlier, our use of RNN dynamics to capture learning-to-learn effects builds directly on pioneering work by Hochreiter and colleagues~\citeyearpar{HochreiterYC01}, in which an RNN model was applied to the problem of function induction~\cite[see also][]{WangEtAl16,santoro2016one}. In contrast to that work, we deployed our RNN as a forward model situated within a larger model-based RL system. In this sense, our implementation bridges between Hochreiter's original proposal and models of motor adaptation that have embedded an adaptive Bayesian model of limb dynamics~\cite[e.g.][]{BernikerK08, GeneweinEtAl15}. The approach we have introduced also relates to other work in which RNNs have been used as forward models in support of motor adaptation, but where multiple fixed models are assumed~\citep{HarunoWK01, PittiEtAl13}, rather than a single adaptive model used here.  These fixed models lack memory, meaning that reweighing fixed models aside, adaptation is only possible by retraining the system.  Implicitly, our work implements a sort of Kalman filter which has also been considered previously in recurrent networks~\citep{WolpertGJ95}. Undertaking a careful comparison between these related approaches and the one we have introduced here offers an important objective for next-step research. 

Our implementation of the reaching task was deliberately minimal, simplifying both the underlying biomechanics and the motor planning process, in order to foreground our central computational proposal. Naturally, a more detailed evaluation of the approach, incorporating a higher degree of empirical constraint, will be desirable in further evaluating the viability of our approach as a theory of motor adaptation. A related opportunity is to consider the potential parallel between the recurrent connectivity underlying the function of our adaptive model and the recurrent connectivity inherent in biological neural circuits underlying motor control and adaptation, including circuits running through the basal ganglia and cerebellum.  

At the same time, however, we feel it may also be fruitful to apply the model-based framework we have introduced here in domains beyond motor control, in particular other domains that display the characteristics of a POMDP and where learning-to-learn effects have been observed. Such tasks are indeed ubiquitous, ranging from structured bandit tasks to video-game play~\citep{WangEtAl16,LakeST15}. To the extent that the framework we have presented here can be adapted and (more challenging) effectively scaled to these other settings, it offers to provide a more general new perspective on the problem of learning-to-learn.  

\subsection{Acknowledgements}
We would like to thank Daniel Braun and Konrad Kording for valuable feedback.

\bibliographystyle{apacite}

\setlength{\bibleftmargin}{.125in}
\setlength{\bibindent}{-\bibleftmargin}

\bibliography{CogSci_Template}

\end{document}